\documentclass[lettersize,journal]{IEEEtran}
\def\eg{\emph{e.g.~}} 
\def\ie{\emph{i.e.~}}

\usepackage[T1]{fontenc} 
\usepackage{comment}
\usepackage{booktabs}
\usepackage{amsmath,amsfonts}
\usepackage{algorithmic}
\usepackage{algorithm}
\usepackage{array}
\usepackage[caption=false,font=normalsize,labelfont=sf,textfont=sf]{subfig}
\usepackage{textcomp}
\usepackage{stfloats}
\usepackage{url}
\usepackage{verbatim}
\usepackage{graphicx}
\usepackage{cite}
\usepackage{caption}
\usepackage{hyperref}
\usepackage[dvipsnames]{xcolor}
\usepackage{orcidlink}

\definecolor{glaucous}{rgb}{0.38, 0.51, 0.71}
\definecolor{stop_grad_red}{rgb}{0.75, 0, 0}

\definecolor{vcrblue}{rgb}{0.02, 0.64, 0.80}
\definecolor{varred}{rgb}{0.93, 0.47, 0.60}
\definecolor{varyellow}{rgb}{0.88, 0.70, 0.47}

\newcommand{\rv}[1]{#1}

\begin{document}

\title{VPNeXt : Rethinking Dense Decoding for \\ Plain Vision Transformer}

\author{Xikai Tang, Ye Huang\orcidlink{0000-0001-5668-5529},~\IEEEmembership{Member,~IEEE, } Guangqiang Yin and Lixin Duan\orcidlink{0000-0002-0723-4016}
\thanks{Xikai Tang is with the School of Information and Software Engineering, University of Electronic Science and Technology of China}
\thanks{
Ye Huang, Guangqiang Yin and Lixin Duan are with the Shenzhen Institute for Advanced Study, University of Electronic Science and Technology of China, 518000 (Ye Huang is the corresponding author)}
}

\markboth{Rethinking Dense Decoding for Plain Vision Transformer}%
{Shell \MakeLowercase{\textit{et al.}}: A Sample Article Using IEEEtran.cls for IEEE Journals}


\maketitle

\begin{abstract}
We present VPNeXt, a new and simple model for the Plain Vision Transformer (ViT). 
Unlike the many related studies that share the same homogeneous paradigms, VPNeXt offers a fresh perspective on dense representation based on ViT.
In more detail, the proposed VPNeXt addressed two concerns about the existing paradigm: (1) Is it necessary to use a complex Transformer Mask Decoder architecture to obtain good representations? (2) Does the Plain ViT really need to depend on the mock pyramid feature for upsampling?
For (1), we investigated the potential underlying reasons that contributed to the effectiveness of the Transformer Decoder and introduced the Visual Context Replay (VCR) to achieve similar effects efficiently.
For (2), we introduced the ViTUp module. 
This module fully utilizes the previously overlooked ViT real pyramid feature to achieve better upsampling results compared to the earlier mock pyramid feature. This represents the first instance of such functionality in the field of semantic segmentation for Plain ViT.
We performed ablation studies on related modules to verify their effectiveness gradually. 
We conducted relevant comparative experiments and visualizations to show that VPNeXt achieved state-of-the-art performance with a simple and effective design.
Moreover, the proposed VPNeXt significantly exceeded the long-established mIoU wall/barrier of the VOC2012 dataset, setting a new state-of-the-art by a large margin, which also stands as the largest improvement since 2015.
The code will be available on \url{https://github.com/edwardyehuang/research/VPNeXt}.
\end{abstract}


\section{Introduction}
\label{sec:intro}

\IEEEPARstart{S}{emantic} 
segmentation is a fundamental computer vision task that classifies the image at the pixel level.
As the most direct way to produce dense representation, semantic segmentation has undergone rapid development over the past decade~\cite{cFCN,cUNet,cPSPNet,cDeepLab,cDeepLabV3Plus,cNonLocal,cDualAttention,cCCNet,cOCR,cCAA,cMaskFormer,cMask2Former,cCAR,cSAR,cCART,cHFGD,cSRRNet,cDeepLabM}.
A high-quality semantic segmentation model can not only benefit numerous application scenarios but also provide strong representations for various downstream computer vision tasks~\cite{cPanopticDeepLab,cDPT,cPanopticFPN,cReELFA}.

Since the introduction of Vision Transformer (ViT)~\cite{cViT} in 2020, numerous researchers have been exploring the use of ViT for visual tasks, including semantic segmentation~\cite{cSegmenter,cSegViT,cRSSeg-ViT}.
In this work, our main focus is the original ViT architecture, also known as Plain Vision Transformer, rather than its variants (e.g. Swin, MaxViT)~\cite{cSwin,cMaxViT,cCPvT}.
The Plain ViT has several advantages because it uses the same architecture as natural language processing (NLP)~tasks, allowing for a smooth transfer of NLP concepts and technologies to visual tasks, such as BEiT~\cite{cBEiT} and MAE~\cite{cMAE}. 
It also helps in creating multimodal models, like the recent Show-O~\cite{cShow-O} and Qwen-VL2~\cite{cQwen-VL,cQwen-VL2}, which combine tokenized images and other modalities into a single Transformer.

\begin{figure}
    \centering
    \includegraphics[width=\linewidth]{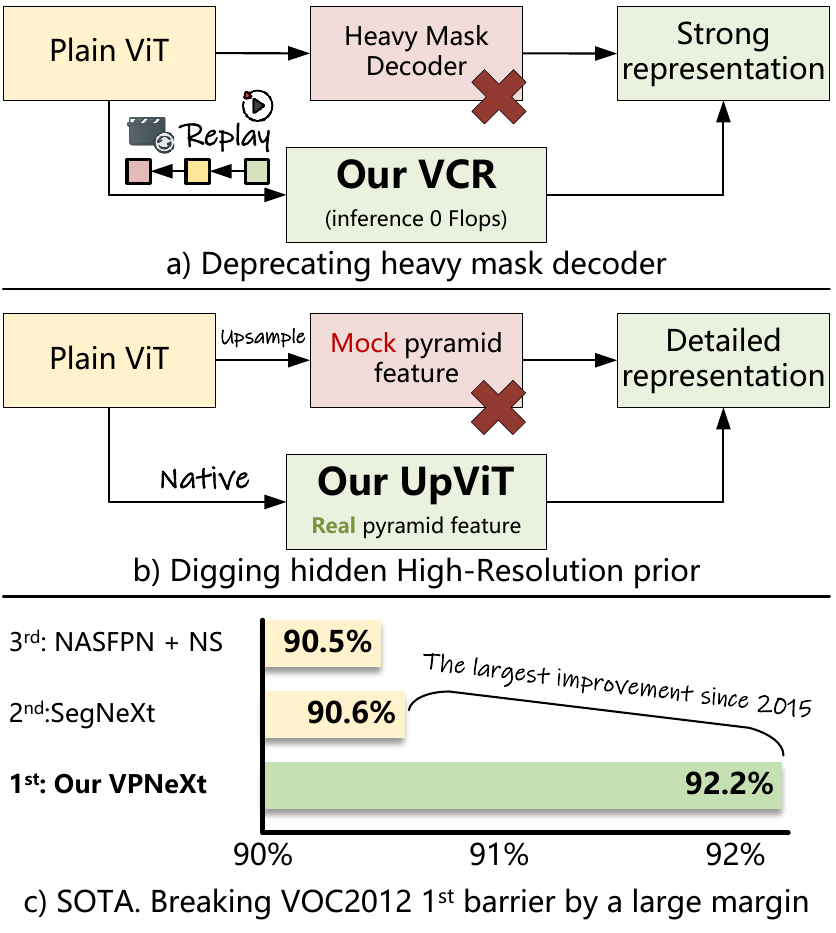}
    \caption{Our main contributions include: 
    1) proposing VCR as an efficient alternative to the computationally intensive Mask Decoder, 
    2) digging hidden native pyramid feature of plain ViT to achieve better upsampling results,
    and 3) breaking the long-standing mIOU wall of the VOC2012~\cite{cPascalVOC} dataset by a large margin, setting a new state-of-the-art with the largest improvement since 2015.}
    \label{fig:abs-1}
\end{figure}

Every coin has two sides.

\rv{\textbf{First}, 
many works on semantic segmentation decoders~\cite{cSegmenter,cMaXDeepLab,cKMaXDeepLab,cMaskFormer,cMask2Former,cSegViT} during the ViT era primarily focused on the DETR~\cite{cDETR} and ViT paradigms. 
Notable examples include Segmenter~\cite{cSegmenter}, MaX-DeepLab~\cite{cMaXDeepLab}, and MaskFormer~\cite{cMaskFormer}, whose decoders all share a similar name: Mask Decoder or Mask Transformer.}
\rv{The strong effectiveness of the mask decoder does not need to be elaborated in this paper, as it has already demonstrated excellent feature regularization capabilities by achieving very high mIOU on multiple benchmark datasets~\cite{cPascalContext,cCocoStuff,cCityScapes}.}

However, the low efficiency of the Mask decoder is a significant concern. 
It is well known that the non-sparse attention operation is global, making its computational efficiency much lower than that of the convolutional-based network. 
The ViT backbone network already contains a large number of attention operations, and the Mask decoder typically requires the addition of three~\cite{cSegmenter,cSegViT} or more attention operations. 
This further decreases overall computational efficiency.

\textbf{Besides}, the plain ViT has an obvious disadvantage in semantic segmentation.
Its tokenizer directly reduces the input image size by at least 16$\times$ times, which is not suitable for semantic segmentation tasks requiring the original size output.
Mature solutions like FPN~\cite{cFPN} and its variants~\cite{cFaPN,cFastFCN,cDeepLabV3Plus,cUper} cannot be implemented because they require multi-resolution pyramid features~\cite{cVGG,cResnet,cXception,cEfficientNet}, which plain ViT obviously cannot provide.
To tackle the ViT's resolution issue, a line of works~\cite{cSETR} opt for transposed convolution or similar methods, which directly upsample feature maps in a stage-wise manner without reference to pyramid features.
However, their impact is often limited as they still depend on the encoder to produce easily upsampling features, which is only slightly better than direct upsampling~\cite{cSegmenter}.
Another line of approaches involves creating a parallel pyramid network to produce high-resolution, low-level features while leveraging the pre-trained features of ViT. 
However, this results in a significant increase in inference overhead.

Therefore, we raise two questions:
\begin{itemize}
    \item Is it necessary to use a complex Transformer Mask Decoder architecture to obtain good representations?
    \item Does the Plain ViT really need to depend on the mock pyramid feature for upsampling?
\end{itemize}

To address these two questions, we propose VPNeXt (ViT context replay and upsample network; 'X' represents new technology.).
%
Specifically, for the first question, VPNeXt includes a novel technology called Visual Context Replay (VCR) to achieve similar effectiveness as Mask-Decoder but with much greater efficiency, as  VCR is only applied during training.
VCR enables the same visual priors to be replayed during the early encoding stages of ViT. 
This allows for the interaction between fine visual priors and early features without increasing computational overhead during the inference stage. 
As a result, this approach leads to improved visual representations.

\rv{In response to the second question, we claim that plain ViT can also effectively extract native pyramid features, similar to those obtained by pyramid networks~\cite{cVGG,cResnet,cEfficientNet, cSwin,cSegFormer,cMaxViT}, as opposed to the mock pseudo-pyramid features derived from resizing high-level features like SETR~\cite{cSETR} and ViTDet~\cite{cViTDet}.
Therefore, we present ViTUp, an effective technique that uncovers high-resolution features previously hidden in ViT and uses them to aid in final upsampling.}

By combining VCR and UpViT, we have successfully developed VPNeXt, a simple, effective, and efficient decoder for ViT.
This approach achieves outstanding performance across multiple benchmark datasets. 

\rv{In summary, our key contributions include :}

\begin{itemize}
    \item We proposed Visual Context Replay (VCR) to achieve similar effectiveness as Mask-Decoder but with much greater efficiency.
    \item We proposed ViTUp to extract native high-resolution features in plain ViT for the pyramid-based upsampling.
    \item The entire solution, VPNeXt, achieves state-of-the-art across multiple benchmark datasets, and also broke the long-standing mIoU wall of the VOC2012 dataset by a large margin, which also stands as the largest improvement since 2015.
\end{itemize}

\section{Related Works}
\label{sec:HFGD:related_work}
\label{sec:UpViT:related_work}

\subsection{Backbones for semantic segmentation}

The primary purpose of backbones is to effectively fit large training sets while enhancing generalization and delivering robust feature representations for downstream tasks (e.g. segmentation, detection).
In general, the challenge of effectively fitting the training set to the deep neural network backbone~\cite{cResnet,cEfficientNet,cViT} has been well addressed through in-network normalization~\cite{cBatchNorm,cLayerNorm,cGroupNorm} and residual connections~\cite{cResnet}.
For generalization and enabling downstream tasks, research on backbones has also made significant progress in the past decade.
For example, the pyramid structure-based backbone network improves generalization by leveraging multi-scale priors. 
It also provides multi-resolution feature maps to downstream tasks, facilitating result upsampling with minimal computational overhead.

However, due to the popularity of natural language processing (NLP) and the research community's interest in unified architectures for multimodal learning, the Vision Transformer (ViT)~\cite{cViT}, which is inspired by the NLP Transformer architecture, has been proposed, with inherent architectural limitations. 
Specifically, it cannot produce pyramid information for downstream decoders to perform multi-scale feature extraction or multi-stage upsampling, which poses significant challenges for semantic segmentation tasks.

Note that, the ViT mentioned here refers specifically to the original plain Vision Transformer (ViT)~\cite{cViT}. 
The ViT pyramid variants~\cite{cSwin,cMaxViT} have fundamentally shifted closer to the CNN architecture, resulting in the loss of some characteristics inherent to the plain ViT, including its unified architectures with NLP.

\subsection{Decoder for semantic segmentation}

The decoder primarily serves the downstream task. 
In semantic segmentation, the decoder usually has two functions: 1) improve the robustness of the encoded features, 2) upsample the feature map back to the original input size.

For the former, i.e. enhancing the robustness of the encoding, common methods include multi-scale feature extraction~\cite{cPSPNet,cDeepLab,cDenseASPP,cFPN} and similarity-based feature extraction (e.g. pixel-wised~\cite{cNonLocal,cDualAttention,cOCNet,cCCNet,cCFNet,cANNN,cCAA} and class-center-wised~\cite{cOCR,cACFNet}). 
Alternatively, this goal can also be achieved by constraining the loss-based regularization~\cite{cCAR,cCPN}.

Recently, the Mask Decoder~\cite{cSegmenter,cMaskFormer,cSegViT,cMask2Former}, inspired by the Transformer decoder, has gained popularity as the leading option for decoders in semantic segmentation. 
It effectively merges the strengths of both similarity-based and class-center-based methods while also integrating the advantages of deep supervision techniques~\cite{cDeeplySupervisedNets,cDeepSupervisedCNN} indirectly.
In the following section, we will further discuss the relationship between deep supervision and the Mask Transformer.
As previously mentioned, while the Mask Transformer is effective, it also brings a significant computational burden, which is one of the issues our work seeks to address.

For upsampling the feature map back to the original input size, the most common approaches are direct upsampling and using pyramid features in a hierarchical manner.

The goal of direct up-sampling is to preserve as much detailed information (i.e. spatial to channel) as possible during the encoding stage while making necessary compromises in the interpolation during up-sampling.
In simple terms, it ensures that the upsampled image's results align with the requirements of the final loss function at full resolution.
One piece of evidence is that even the Segmenter~\cite{cSegmenter}, which upsampled directly at a 1/16 resolution, can produce good detailed results.
Nevertheless, as previously stated, encoders must make compromises in detail interpolation.

Using an upsampler can effectively reduce the load on the encoder. 
Although existing upsamplers still have many issues~\cite{cHFGD}, they have proven to be very effective in numerous studies~\cite{cUNet,cFPN,cFaPN,cCART,cFastFCN}.
Among the various upsamplers, the one based on pyramid information is the most typical and widely utilized.
Unfortunately, the current popular backbone in the research community, plain ViT~\cite{cViT}, is unable to provide multi-stage pyramid information.

Several downstream ViT-based research efforts, including those for semantic segmentation~\cite{cSETR}, attempt to forcefully apply pyramid upsamplers to ViT.
Their common method is to upsample intermediate features~\cite{cSETR} of ViT or directly upscale the final high-level features~\cite{cViTDet} to various scales, mocking pyramid features.
These methods offer only a slight advantage in extracting multi-scale features and are essentially no different from direct upsampling~\cite{cSegmenter} when it comes to providing high-resolution features.
This occurs because the intermediate or final features lack the native high-resolution details, relying instead on the hope that some lost spatial information remains preserved in the channels.

In this work, we revisit the architecture design of the ViT. 
The decoder we propose is efficient and lightweight while effectively mining native ViT high-resolution pyramid features, aiding in the efficient feature upsampling.

\section{Preliminary}

Deep supervision techniques~\cite{cDeeplySupervisedNets,cDeepSupervisedCNN} have proven to be effective in experiments over the past decade.
Using auxiliary loss~\cite{cPSPNet,cDualAttention,cCAA} on the backbone is a common practice in deep supervision.
Although there is no theoretical consensus on its effectiveness, we believe that deep supervision helps the early shallow layers of the network align better with the optimization target (i.e. loss function).

One of the reasons why masked decoders are so effective is that there is a good chance that they are indirectly using deep supervision or even an enhanced version of it.
In the mask decoders~\cite{cMaXDeepLab,cKMaXDeepLab,cMask2Former}, the class token interacts repeatedly with feature maps from various levels, including both deep and shallow layers, through cross-attention. 
This interaction not only ensures that the shallow layers can indirectly enjoy the deep supervision from the optimization target but also helps them align more effectively with the class token, which is essential for the final classification.

\section{Proposed Method}
\label{sec:method}
This work presents two methods. 
The first is Visual Context Replay (VCR), which is a simple, efficient, and effective technique for enhancing the decoder's input features. 
The second is UpViT, which reveals the inherent high-resolution features that are typically thought to be absent in ViT~\cite{cViT}.

\subsection{Visual Context Replay (VCR)}

\begin{figure*}[t]
    \centering
    \includegraphics[width=1.0\linewidth]{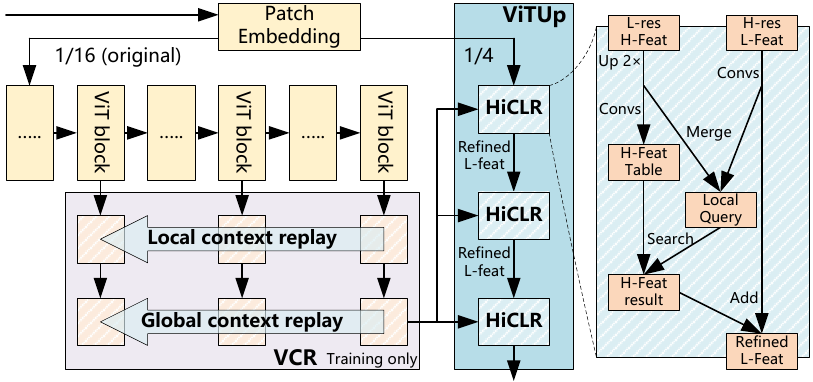}
    \caption{Our proposed VPNeXt consists of two main modules: VCR and ViTUp, which focus on enhancing features efficiently and addressing upsampling challenges, respectively.}
    \label{fig:overall}
\end{figure*}

VCR is a lightweight and innovative feature enhancement technology for deep supervision that performs comparably to the mask decoder while having zero inference overhead.

For an N-layer ViT backbone network, we define the output feature map of each layer as $\mathbf{x}_{i}$, $i$ represents the layer index.
Improving the robustness of those intermediate layers can be helpful for the final output.
In mask decoders, the outputs of two to three intermediate layers are typically optimized using skip connections. 
At VCR, we also optimize two intermediate layers, referred to as $\mathbf{x}_{a}$ and $\mathbf{x}_{b}$.

To avoid the inference computation overhead of the Mask decoder's attention while still providing supervision for these intermediate layers, traditional deep supervision appears to be a suitable approach without introducing extra inference computation overhead.
However, as mentioned in previous sections, deep supervision supervises each intermediate layer independently, which does not align with others using class (mask) tokens as the mask decoder does, resulting in less effectiveness.

One simple improvement idea is to utilize the final output feature map from the last layer (which we define as $\mathbf{x}_{z}$) to supervise the feature maps of the intermediate layers, in order to enforce the alignment.
We refer to this strategy as "naive align," which has the following loss function:
\begin{equation}
    L_{\text{naive-align}} = \sum_{i\in \{ a,b\}}\text{MSE}(\mathbf{x}_{z},~\mathbf{x}_{i})
\end{equation}

The alignment targets the last layer feature map because it typically provides a better representation than intermediate layers.
However, this operation is clearly ineffective. 
It is impossible to generate the same representation in an early intermediate layer as in the final layer because each intermediate layer serves a specific function that contributes to the progressive creation of the final output. 
If this were not true, we would only need a single layer without hidden layers.

Thus, We need to have control over what is effective for alignment.
This is what VCR represents. 
The replay mechanism takes the visual context from the final layer and replays it to the intermediate layers. 
This helps the deep supervision achieve the same level of effectiveness as the mask decoder.

Specifically, we believe only the spatial relation is worthwhile and useful for the alignment because the spatial context-aggregation (e.g., convolution, spatial attention, and MLP-mixer) is essential when performing the pixel-encoding.
Therefore, VCR aligns the intermediate layers towards two dimensions: local context and global context, to achieve alignment at both short-range and long-range levels.

\subsubsection{Local context replay}
The implementation of local context replay utilizes a deformable convolutional operation, where the "offset" serves as the key parameter for calculating the positions relative to the center for local context aggregation.
The local context replay operation is illustrated below:
\begin{equation}
    \mathbf{\gamma}_{i} = \text{Deformable}(\mathbf{x}_{i}~,~\mathbf{\sigma}_{z},~\mathbf{\varrho_{i}}),~i \in \{a, b\}
\end{equation}

During the replay process, the learnable offset $\mathbf{\sigma}_{z}$ from the final output layer $z$ will be synchronized directly with the intermediate layers. 
Notation $\mathbf{\varrho_{i}}$ stands for the other learnable parameters in deformable convolution.
The replay mechanism based alignment allows the intermediate layers to perform local context aggregation operations at the same position.

\subsubsection{Global context replay}
Unlike the positional-sensitivity local context replay, the global context replay, which is the final step of VCR, emphasizes the context most relevant to classification. 
The global context replay is based on the concept that intra-class pixels within the same context share similar feature representations.
As stated before, the intermediate layers struggle to learn strong feature representations related to specific categories since they serve their own purpose.
We can allow them to interact only with intra-class pixels, enabling them to learn the useful encoding process effectively.
To accomplish this, VCR replays the dot-product pixel affinity $\mathbf{\Lambda}_{z}$ as a spatial relation prior and regularizes the encoding of intermediate layers.
The global context replay operation is illustrated below:

\begin{equation}
    \mathbf{y}_{i} = \text{Attention}(\mathbf{\Lambda}_{z},~\phi_{i}(\mathbf{\gamma}_{i})),~i\in \{a, b\}
\end{equation}

The notation $\phi_{i}$ represents the linear projection of the attention operation. Its input $\mathbf{\gamma}_{i}$ is the 'value' derived from the previous equation.\\

After performing the VCR, which includes both local context replay and global context replay, the feature representations from the intermediate layers align effectively with the final feature representation. This alignment is achieved without any additional inference overhead, serving as a form of deep supervision. Furthermore, the VCR method demonstrated performance comparable to that of mask decoders in our experiments. 
We will provide the details of these experiments in the following sections.

\subsection{ViTUp}

As described in the earlier sections, ViT does not generate multi-scale (\eg different resolution) pyramid feature maps across multiple stages, necessitating the creation of mock multi-scale pyramid feature maps before utilizing commonly applied pyramid upsamplers, which renders them ineffective (please refer to the previous sections for more details).

However, we observed that the plain ViT generates a hidden high-resolution pyramid feature map, which can be effectively utilized for pyramid upsampling.
Given an input image $\mathbf{I}$, the plain ViT model uses patch embedding for tokenization, resulting in a feature map $\mathbf{x}_{0}$ that is typically 1/16 the size of the input image, as shown below:

\begin{equation}
    \mathbf{x}_{0} = \theta(\mathbf{I}, \mathbf{K}_{\text{16}}, \mathbf{S}_{\text{16}})
    \label{eq::patch-embedding}
\end{equation}

In practical applications, patch embedding is implemented using a 2D convolution operation $\theta$. 
In this process, both the kernel size $\mathbf{K}$ and stride $\mathbf{S}$ are set to the same value as the patch size. 
For example, if the patch size is 16, the kernel size and stride are also set to 16, as represented by $\mathbf{K}_{\text{16}}$ and $\mathbf{S}_{\text{16}}$ in equation~\ref{eq::patch-embedding}.

Inspired by the DeepLab series~\cite{cDeepLab}, reducing or eliminating the stride enables the creation of a larger resolution feature map without affecting the range of spatial context aggregation, also known as the receptive field. 
By adjusting the stride of patch embedding to a value smaller than the patch size (for example, using a stride of 4 for the typical pyramid upsampler), a hidden high-resolution pyramid feature map with a large size can be extracted, as shown below.

\begin{equation}
    \mathbf{x}_{0} = \theta(\mathbf{I}, \mathbf{K}_{\text{16}}, \mathbf{S}_{\text{4}})
\end{equation}

Note that, in VPNeXt, we only need to calculate the patch embedding $\theta$ once for 1/4, then downsample it by a factor of 4 for 1/16 required by the ViT.

After obtaining the hidden high-resolution pyramid feature map, we observed a minor difference compared to existing pyramid upsampler-based models. 
For instance, models like FPN, UperNet, and FaPN typically utilize multi-scale (two high-resolution) feature maps (excluding the final output from the backbone), while our approach extracts only a single high-resolution feature map. 
In comparison to DeepLabV3+, although it also uses a single high-resolution feature map, this feature map is derived from a deeper intermediate layer of the backbone than ours. 
The feature map from this deeper intermediate layer generally offers better encoding and has a smaller alignment gap with the backbone's final output.

To enhance the effectiveness and smoothness of the pyramid upsampler with our shallow pyramid feature map, we proposed the High-Level Context Local Refiner (HiCLR).
HiCLR employs a coarse-to-fine strategy that uses multiple iterations of refinement to progressively reduce the alignment gap between the high-resolution pyramid feature from the shallow layer and the low-resolution feature from the final backbone output.
Each refinement iteration takes two inputs: a high-level backbone feature map and an upsampled feature map. In the first iteration, the upsampled feature map is derived from the extracted hidden high-resolution pyramid feature. 
In subsequent iterations, the upsampled feature map is obtained from the output of the previous iteration.

In the refinement process, we use a method similar to VCR, where the spatial context of high-level features is leveraged to align the upsampled features. The key difference is that HiCLR concentrates exclusively on local context refinement, as restoring the missing local details is sufficient for the upsampling operations. 
This concept is also widely implemented in other upsamplers, such as the $3\times3$ convolutions used in DeepLab V3+ and UperNet.
\section{Training details}
\label{sec:HFGD:training_settings}

Unless specified otherwise, the training settings for our proposed VPNeXt are similar to existing works that use ViT mask decoders~\cite{cSETR,cSegViT,cMask2Former}.
This includes the AdamW optimizer, a batch size of 16, and the use of clipnorm along with a mask loss that combines focal and dice losses.

Given that this work focuses exclusively on the plain ViT backbone, all the experiments we conducted are based on the plain ViT without pyramid modifications. 
Following common practices, the weights of the ViT are initialized through modern pre-training~\cite{cAugReg,cEVA}.

To accommodate new readers in the field, we utilize the commonly used Mean Intersection over Union (mIOU) metric to evaluate the prediction accuracy of our model.
\section{Experiments on Pascal Context Dataset}
The Pascal Context~\cite{cPascalContext} dataset comprises 4,998 training images and 5,105 testing images. We utilize its 59 semantic classes to perform ablation studies and experiments, following common practice. Unless otherwise specified, we train the models on the training set for 20K iterations.

In the ablation studies, we follow the VPNeXt's forward propagation sequence. 
First, we assess the effectiveness of VCR alone, and then we incorporate ViTUp to evaluate its ability to upsample the feature maps produced by VCR.
Finally, we conducted an analysis of computational overhead to evaluate the efficiency of our proposed VPNeXt.


\subsection{Ablation studies on VCR}
We compare our proposed VCR with a mask decoder (w/o pyramid, \eg segmenter~\cite{cSegmenter}) and deep supervision, as discussed in previous sections. 
As shown in Table~\ref{tab:exps:vcr-ablation-studes}, incorporating visual context in deep supervision results in an even better mIOU than the mask decoder (68.83\% vs 67.88\%). 

Additionally, we conducted ablation studies to determine the optimal number of deep supervision layers to use. 
The results in Table~\ref{tab:exps:vcr-ablation-studes} indicate that the mIOU reaches its highest value when two intermediate layers are employed for VCR-oriented deep supervision.

\begin{table}[ht]
    \centering
    \small
    \caption{Ablation studies on VCR, all the results are obtained under single-scale without flipping.
    All baseline models are trained using the same backbone and settings.
    \textit{DS:} Deep supervision.
    }
    \resizebox{\linewidth}{!}{
    \begin{tabular}{l|c|c}
       \toprule
       Methods \quad & Num\# DS layers &  \quad mIOU(\%)\quad \\
       \midrule
       Deep supervision & 2 & 66.50 \\
       \midrule
       Mask decoder  w/o pyramid & 2 (implicit) & 67.88  \\
       \midrule
       Our VCR & 1 & 68.43\\
        & \textbf{2} & \textbf{68.83}\\
        & 3 & 68.56 \\
       \bottomrule
    \end{tabular}
    }
    \label{tab:exps:vcr-ablation-studes}
\end{table}

\begin{table}[ht]
    \centering
    \caption{Ablation studies on ViTUp, all the results are obtained under the single-scale without flipping.
    All baseline models are trained using the same backbone and settings.
    }
    \resizebox{\linewidth}{!}{
    \begin{tabular}{l|c|c}
       \toprule
       Methods & Num\# HiCLR layers&  mIOU(\%) \\
       \midrule
       Bilinear  & 0 & 68.83 \\
       \midrule
       Mock pyramid & 2 & 69.01 \\
       \midrule
       Our real pyramid \quad & 1 & 69.50 \\
        & 2 & 69.87 \\
        & \textbf{3} & \textbf{70.00}\\
        & 4 & 69.81 \\
        & 5 & 69.43 \\
       \bottomrule
    \end{tabular}
    }
    \label{tab:exps:ViTUp-ablation-studes}
\end{table}

\begin{table}[ht]
    \centering
    \caption{
    Computational cost analysis for VPNeXt.
    All baseline models use the same backbone and settings.
    }
    \resizebox{\linewidth}{!}{
    \begin{tabular}{l|c|c}
       \toprule
       Methods    &  
       Pyramid Upsampler \quad &
       \quad GFlops \quad \quad  \\
       \midrule
       Deep supervision  \quad & - & 356.69\\
       \midrule
       Mask decoder & & 359.99  \\
       & \checkmark & > 2000 \\
       \midrule
       Our VPNeXt & & 356.69 \\
        & \checkmark & 552.85 \\
       \bottomrule
    \end{tabular}
    }
    \label{tab:exps:cost-ablation-studes}
\end{table}

\begin{table}[ht!]
\centering
\small
\caption{
Comparisons to state-of-the-art methods on Pascal Context dataset.
\textit{SS}: Single-scale performance w/o flipping.
\textit{MF}: Multi-scale performance w/ flipping.
``-'' in column \textit{SS} indicates that this result was not reported in the original paper.
}
\resizebox{\linewidth}{!}
{\def\arraystretch{1} \tabcolsep=0.55em 
\begin{tabular}{l|c|c|c|c}
\toprule
Methods & Backbone & Avenue &\multicolumn{2}{c}{mIOU(\%)} \\
& & & SS & MF \\
\midrule
\midrule
SETR~\cite{cSETR}           & ViT-L           & CVPR'21 & 55.0 & 55.8 \\
DPT~\cite{cDPT}             & ViT-Hybrid      & ICCV'21 & - & 60.5 \\
OCNet~\cite{cOCNet}         & HRNet-W48       & IJCV'21 & - & 56.2 \\
CAA~\cite{cCAA}             & EfficientNet-B7 & AAAI'22 & 59.4 & 60.5 \\
CAA + CAR~\cite{cCAR}        & ConvNeXt-L      & ECCV'22 & 62.7 & 63.9 \\
SegNeXt~\cite{cSegNeXt}     & MSCAN-L         & NIPS'22 & 59.2 & 60.9 \\
SegViT~\cite{cSegViT}       & ViT-L            & NIPS'22 & 64.1 & 65.3 \\
SenFormer~\cite{cSenFormer}  & Swin-L          & BMVC'22 & 63.1 & 64.5\\
TSG~\cite{cTSG}             & Swin-L           & CVPR'23 & - & 63.3 \\
IDRNet~\cite{cIDRNet}        & Swin-L           & NIPS'23 & 63.8 & 64.5 \\
APPNet~\cite{cAPPNet}       &SenFormer-L       & TCSVT'23 & - & 63.7 \\
ViT-Adapter-L~\cite{cViTAdapter} & ViT-L       & ICLR'23 & 67.8 & 68.2 \\
InternImage~\cite{cInternImage} & InternImage-H & CVPR'23 & 69.7 & 70.3 \\
SegViTv2~\cite{cSegViTv2}       & ViT-L          & IJCV'23 & - & 67.1 \\
PFT~\cite{cPFT}             & ResNet-101       & TMM'24 & 55.2 & 57.3 \\
CART~\cite{cCART}           &EfficientNet-L2   & TCSVT'24 & 66.0 & 67.5 \\
HFGD~\cite{cHFGD}           &ConvNeXt-L        & TCSVT'24 & 64.9 & 65.6 \\
ALGM~\cite{cALGM}           &ViT-L             & CVPR'24 & - & 58.0 \\
SILC~\cite{cSILC}           &SILC-C-L          & ECCV'24 & - & 61.5 \\
\midrule
VPNeXt (w/o ViTUp)         & ViT-L & - & \textbf{68.8} & \textbf{69.7} \\
VPNeXt          & ViT-L & - & \textbf{70.0} & \textbf{71.1}  \\
\bottomrule
\end{tabular}
}
\label{tab:SOTA-PascalContext}
\end{table}

\subsection{Ablation studies on ViTUp}
We then assess the mIOU of our proposed ViTUp. 
As shown in Table~\ref{tab:exps:ViTUp-ablation-studes}, the real pyramid feature provided by our ViTUp, enhanced by HiCLR, reached 69.50\% mIOU, significantly outperforms both bilinear interpolation and mock pyramids (69.50\% vs 68.83\% vs 69.01\%). 
Furthermore, applying refinement three times yields 70.00\% mIoU, making it the best ViTUp configuration for VPNeXt.

\subsection{Computational cost analysis}
To demonstrate the high efficiency of VPNeXt, we conducted a computational analysis on two setups: VCR (VPNeXt w/o pyramid upsampler) and the complete VPNeXt with ViTUp.
For fair comparisons, we utilized Segmenter~\cite{cSegmenter} as the Mask decoder w/o a pyramid upsampler, and Mask2Former-based~\cite{cMask2Former} Vit-adapter~\cite{cViTAdapter} and PlainSeg~\cite{cPlainSeg} as Mask decoders with/a pyramid upsampler.

Table~\ref{tab:exps:cost-ablation-studes} shows that VCR and deep supervision have the same Flops, indicating that VCR provides high-quality representations without adding any computational overhead (see previous subsections for details).
Table~\ref{tab:exps:cost-ablation-studes} also shows that ViTUp delivers high-resolution pyramid features and strong mIoU while having significantly lower computational overhead compared to previous mask decoders that rely on mock pyramid features.

\subsection{Compare with state-of-the-arts}
To fully showcase the performance superiority of VPNeXt, we compared it with state-of-the-art methods on the Pascal Context dataset.
Note that, only methods published by the time this paper was completed can be compared.
As shown in Table~\ref{tab:SOTA-PascalContext}, our proposed VPNeXt significantly outperforms the compared methods, including the previous state-of-the-art techniques ViT-Adapter and InternImage. 
Moreover, even without using ViTUp (\ie with only VCR), VPNeXt still outperforms most methods.

\begin{table}[th!]
\centering
\small
\caption{
Comparisons to state-of-the-art methods on COCOStuff164k dataset.
\textit{SS}: Single-scale performance w/o flipping.
\textit{MF}: Multi-scale performance w/ flipping.
}
\resizebox{\linewidth}{!}
{\def\arraystretch{1} \tabcolsep=0.55em 
\begin{tabular}{l|c|c|c|c}
\toprule
Methods & Backbone  & Avenue &\multicolumn{2}{c}{mIOU(\%)}\\
& & & SS & MF \\
\midrule
\midrule
OCR~\cite{cOCR,cHRFormer} & HRFormer-B & NIPS'21 & - & 43.3 \\
SegFormer~\cite{cSegFormer} & MiT-B5 & NIPS'21 & - & 46.7 \\
CAA~\cite{cCAA} & EfficientNet-B5 & AAAI'22 & - & 47.3 \\
SegNeXt~\cite{cSegNeXt} & MSCAN-L & NIPS'22 & 46.5 & 47.2 \\
RankSeg~\cite{cRankSeg} & ViT-L & ECCV'22 & 46.7 & 47.9 \\
ViT-Adapter~\cite{cViTAdapter} & ViT-L & ICLR'23 & - & 52.0 \\
InternImage~\cite{cInternImage} &  InternImage-H & CVPR'23 & 52.6 & - \\
CART~\cite{cCART} & EfficientNet-L2 & TCSVT'24 & 50.2 & 50.9 \\
HFGD~\cite{cHFGD} & ConvNeXt-L & TCSVT'24 & 49.0 & 49.4 \\
ALGM~\cite{cALGM} & ViT-L & CVPR'24 & - & 47.4\\
\midrule
VPNeXt & ViT-L & - & \textbf{53.0} & \textbf{53.7} \\
\bottomrule
\end{tabular}
}
\label{tab:SOTA-COCOStuff164k}
\end{table}

\section{Experiments on COCOStuff164k Dataset}

COCOStuff164k has become increasingly popular in recent years and poses a significant challenge for semantic segmentation models due to its high diversity, consisting of 118,000 training images and 5,000 testing images, along with its complexity of 171 classes.

In Table~\ref{tab:SOTA-COCOStuff164k}, our VPNeXt model outperforms previous state-of-the-art methods, including ViT-Adapter and InternImage, by a significant margin.

\section{Experiments on Cityscapes Dataset}

Cityscapes is a semantic segmentation dataset featuring high-resolution images of road scenes with precise annotations. 
It includes 19 labeled classes and contains 2,975 training images and 500 validation images.
We only compare methods trained on the Cityscapes fine annotations, similar to many other works.~\cite{cSegFormer,cKMaXDeepLab}.

As shown in Table~\ref{tab:SOTA-Cityscapes}, our proposed VPNeXt, leveraging ViTUp's strong capabilities, performs comparably to state-of-the-art pyramid-based models (e.g., HFGD~\cite{cHFGD} and DPP~\cite{cDDP}) on high-resolution images.

\begin{table}[ht]
\centering
\small
\caption{
Comparisons to state-of-the-art methods on Cityscapes validation set.
\textit{SS}: Single scale performance w/o flipping.
\textit{MF}: Multi-scale performance w/ flipping.
}
\resizebox{\linewidth}{!}{
\begin{tabular}{l|c|c|c|c}
\toprule
Methods &Backbone & Avenue &\multicolumn{2}{c}{mIOU(\%)}\\
&  & & SS & MF \\
\midrule
\midrule
RepVGG\cite{cRepVGG} & RepVGG-B2 & CVPR'21 & - & 80.6 \\
SETR~\cite{cSegFormer} &ViT-L & CVPR'21 & 79.2 & 81.0 \\
Segmenter~\cite{cSegmenter} &ViT-L & ICCV'21 & 79.1 & 81.3 \\
OCR~\cite{cOCR,cHRFormer} &HRFormer-B & NIPS'21 & 81.9 & 82.6 \\
HRViT-b3~\cite{cHRViT}  & MiT-B3 & CVPR'22 & - & 83.2\\
FAN-L~\cite{cFANs} & FAN-Hybrid & ICML'22 & - & 82.3 \\
SegDeformer~\cite{cSegDeformer} & Swin-L & ECCV'22 & - & 83.5 \\
GSS-FT-W~\cite{cGSS} & Swin-L & CVPR'23 & 80.5 & - \\
TSG~\cite{cTSG} & Swin-L & CVPR'23 & - & 83.1 \\
STL~\cite{cSTL} & FAN-Hybrid & ICCV'23 & - & 82.8 \\
DDP(Step 3)~\cite{cDDP} & ConvNeXt-L & ICCV'23 & 83.2 & 83.9 \\
StructToken~\cite{cStructToken} & ViT-L & TCSVT'23 & 80.1 & 82.1 \\
GSCNN(EPL)~\cite{cEPL} & WRNet-38 & TMM'23 & - & 81.8\\
TFRNet~\cite{cTFRNet} & ViT-L & TMM'24 & - & 82.9\\
CART~\cite{cCART} & ConvNeXt-L & TCSVT'24 & 82.8 & 83.6\\
HFGD~\cite{cHFGD} & ConvNeXt-L & TCSVT'24 & 83.2 & 84.0 \\
ALGM~\cite{cALGM} & ViT-L & CVPR'24 & - & 79.5 \\
\midrule
VPNeXt   & ViT-L              & - & 83.0 & \textbf{84.4} \\
\bottomrule
\end{tabular}
}
\label{tab:SOTA-Cityscapes}
\end{table}

\section{Experiments on VOC2012}
VOC2012 is one of the most classic datasets of semantic segmentation. 
It features a small number of categories (21 w/ background), medium resolution, and high annotation accuracy, which allowed earlier methods to achieve a mIoU of 89\% between 2018 and 2019. 

In subsequent years, although stronger methods were developed, they only resulted in slight improvements to the mIoU—usually by a few tenths (\ie < 0.5\%). 
Eventually, SegNeXt~\cite{cSegNeXt} raised the mIoU to 90.6\% in 2022, and since then, no other method has surpassed this mIOU wall. 
As a result, SegNeXt was considered the ceiling for this dataset.

\begin{table}[h]
    \normalsize
    \centering
    \begin{tabular}{cr}
        \midrule
          \quad there is no wall \quad\\
         &\textit{\quad\quad-- Sam Altman~\cite{cNoWall}}\\
         \midrule
    \end{tabular}
\end{table}

Today, our proposed VPNeXt has broken this wall.
As shown in Table.~\ref{tab:SOTA-VOC2012},
in terms of mIoU, our proposed VPNeXt not only outperforms SegNeXt but also exceeds SegNeXt by nearly 2\%., which also stands as the largest improvement since 2015.
Remarkably, VPNeXt excels in long-tailed categories (\eg chair, monitor) that have traditionally posed challenges for nearly all prior methods.

\rv{It is worth noting that the training strategies employed, including data augmentation techniques and the use of the combined VOC2012~\cite{cPascalVOC} and SBD~\cite{cSBD} datasets (also known as `trainaug'), are consistent with those used in existing works, such as DeepLabV3+~\cite{cDeepLabV3}.}

\begin{table*}[th!]
\centering
\small
\caption{
New breakthroughs in the VOC2012 leaderboard!
Due to limited space on the page, we have simplified some category names (e.g., "Aero Plane" to "Plane") and only listed the top \rv{15} methods. 
Zoom in to see better.
To view the full leaderboard, please visit \url{http://host.robots.ox.ac.uk:8080/leaderboard/displaylb_main.php?challengeid=11&compid=6}. (The main website might be down, see web archive mirror: \url{https://web.archive.org/web/20250620175614/http://host.robots.ox.ac.uk:8080/leaderboard/displaylb_main.php?challengeid=11&compid=6})
}
\resizebox{\linewidth}{!}
{\def\arraystretch{1} \tabcolsep=0.55em 
\begin{tabular}{l|c|cccccccccccccccccccc}
\toprule
Methods & 
Mean & 
Plane &
Bicycle &
Bird &
Boat &
Bottle &
Bus &
Car &
Cat &
Chair &
Cow &
Table &
Dog &
Horse &
Motor &
Person &
Plant &
Sheep &
Sofa &
Train &
Monitor \\
\midrule
\midrule
\textbf{Our VPNeXt} & \textbf{92.2}& 98.9 & 78.5 & 98.6 & 92.1 & 92.3 & 95.2 & 96.8 & 96.1 & 70.7 & 98.8 & 79.9 & 96.0 & 98.4 & 96.9 & 95.8 & 89.8 & 98.2& 78.1 & 96.6 & 91.3\\
\midrule
\midrule
SegNeXt& 90.6 & 98.3 & 85.0 & 97.6 & 88.3 & 91.3 & 97.5 & 91.4 & 98.3 & 60.4 & 96.7 & 85.0 & 95.7 &98.2 & 94.2 &92.7 &82.5 & 97.3 & 77.7 & 93.1 & 84.3\\
\midrule
NAS-FPN(NS)& 90.5 & 98.0 & 84.8 & 89.6 & 88.2 & 91.0 & 98.3 & 93.0 & 98.5 & 57.5& 98.4 & 81.8 & 98.4 & 98.0 & 95.8 & 93.2 & 83.2 & 97.8 & 75.0 & 91.8 & 90.0\\
\midrule
DeepLabv3+(JFT) & 89.0 & 97.5 & 77.9 & 96.2 & 80.4 & 90.8 & 98.3 & 95.5 & 97.6 & 58.8 & 96.1 & 79.2 & 95.0 &97.3 & 94.1 & 93.8 & 78.5 & 95.5 & 74.4 & 93.8 & 81.6\\
\midrule
RecoNet152 & 89.0 & 97.3 & 80.4 & 96.5 & 83.8 & 89.5 & 97.6 & 95.4 & 97.7 & 50.1 &	96.8 & 82.6 &	95.1 & 97.7 &	95.1 & 92.6 & 80.2 & 95.2 & 71.7 & 92.1 & 83.8\\
\midrule
AASPP & 88.5 & 97.4 & 80.3 & 97.1 & 80.1 & 89.3 & 97.4 & 94.1 & 96.9 & 61.9 & 95.1 & 77.2 & 94.2 & 97.5 & 94.4 & 93.0 & 72.4 & 93.8 & 72.6 & 93.3 & 83.3\\
\midrule
SRC-B & 88.5 & 97.2 & 78.6 & 97.1 & 80.6 & 89.7 & 97.4 & 93.7 & 96.7 & 59.1 & 95.4 & 81.1 & 93.2 & 97.5 & 94.2 & 92.9 & 73.5 & 93.3 & 74.2 & 91.0 & 85.0\\
\midrule
SepaNet & 88.3 & 97.2 & 80.2 & 96.2 & 80.0 & 89.2 & 97.3 & 94.7 & 97.7 & 48.6 & 95.0 & 81.6 & 95.2 & 97.5 & 95.1 & 92.7 & 79.5 & 95.4 & 68.8 & 90.9 & 83.4\\
\midrule
EMANet152 & 88.2 & 96.8 & 79.4 & 96.0 & 83.6 & 88.1 & 97.1 & 95.0 & 96.6 & 49.4 & 95.4 & 77.8 & 94.8 & 96.8 & 95.1 & 92.0 & 79.3 & 95.9 & 68.5 & 91.7 & 85.6\\
\midrule
KSAC-H & 88.1 & 97.2 & 79.9 & 96.3 & 76.5 & 86.5 & 97.5 & 94.5 & 96.9 & 54.8 & 95.3 & 81.4 & 93.7 & 97.2 & 94.0 & 92.8 & 77.3 & 94.4 & 73.5 & 91.1 & 83.4\\
\midrule
SpDConv2 & 88.1 & 96.9 & 79.7 & 96.8 & 80.2 & 87.8 & 98.0 & 92.3 & 96.0 & 57.2 & 95.8 & 82.1 & 92.3 & 97.3 & 93.6 & 93.0 & 71.4 & 92.3 & 75.8 & 90.7 & 83.8\\
\midrule
FillIn & 88.0 & 97.1 & 80.8 & 96.7 & 77.6 & 89.2 & 97.4 & 92.2 & 96.9 & 58.3 & 94.3 & 79.4 & 93.1 & 97.3 & 94.4 & 93.2 & 73.6 & 93.0 & 72.6 & 89.7 & 83.4\\	
\midrule
MSCI & 88.0 & 96.8 & 76.8 & 97.0 & 80.6 & 89.3 & 97.4 & 93.8 & 97.1 & 56.7 & 94.3 & 78.3 & 93.5 & 97.1 & 94.0 & 92.8 & 72.3 & 92.6 & 73.6 & 90.8 & 85.4\\
\midrule
ExFuse & 87.9 & 96.8 & 80.3 & 97.0 & 82.5 & 87.8 & 96.3 & 92.6 & 96.4 & 53.3 & 94.3 & 78.4 & 94.1 & 94.9 & 91.6 & 92.3 & 81.7 & 94.8 & 70.3 & 90.1 & 83.8\\
\bottomrule
\end{tabular}
}
\label{tab:SOTA-VOC2012}
\end{table*}
\section{Visualization}

\begin{figure}[ht!]
    \centering
    \includegraphics[width=\linewidth]{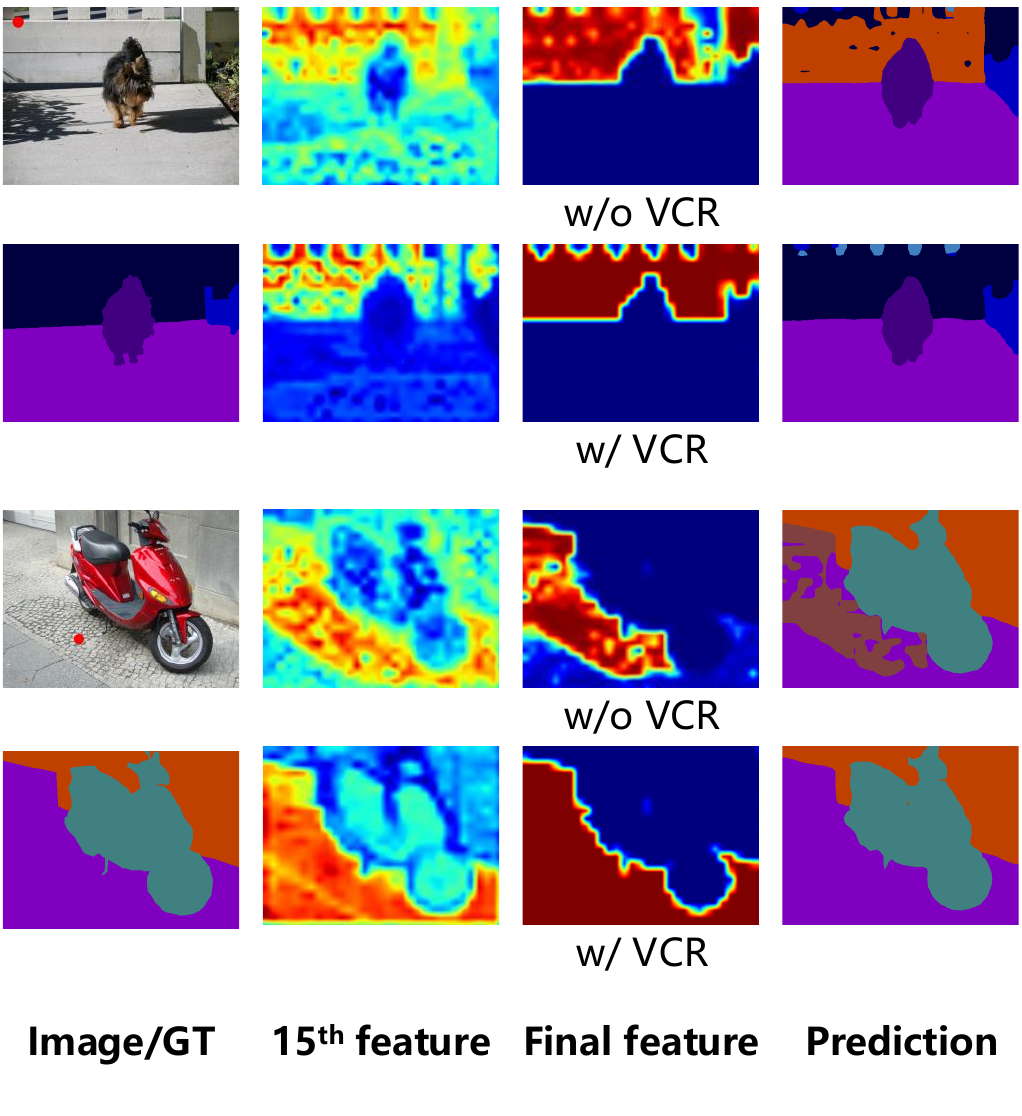}
    \caption{\rv{Visualization analysis of the intermediate feature map for the VCR, based on the 15th layer of the ViT on the Pascal Context dataset, at the pixel position marked by the red dot, the replay-optimized feature map displays significantly stronger and more detailed semantic information concerning intra-class pixels.
    Note that, in order to maintain controlled variables, modules other than VCR, such as UpViT are not utilized.}}
    \label{fig:vcr_feature_vis}
\end{figure}

\begin{figure}[ht!]
    \centering
    \includegraphics[width=1.0\linewidth]{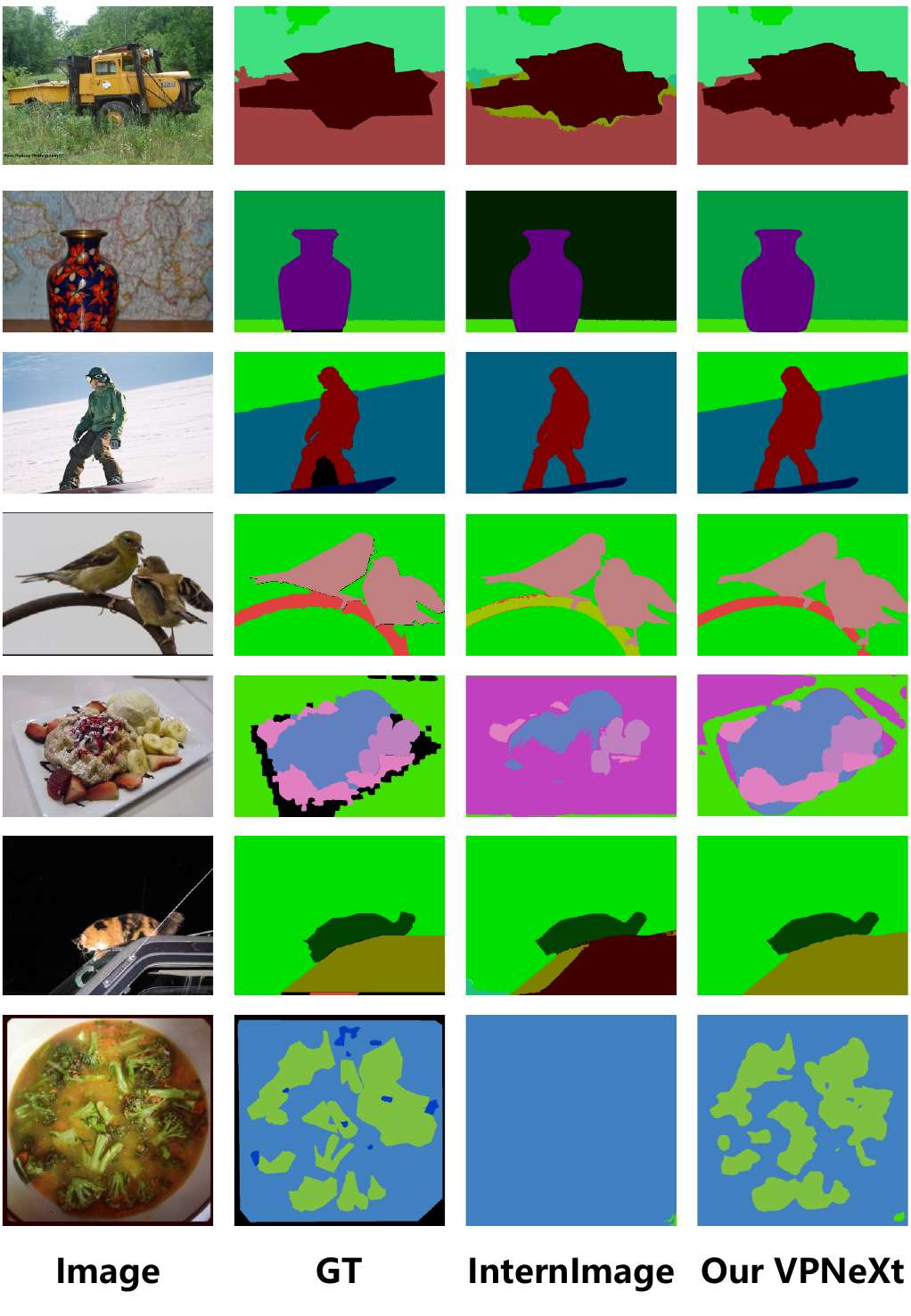}
    \caption{Visual comparison between Mask2Former + InternImage-H~\cite{cMask2Former,cInternImage} and our proposed VPNeXt on the COCOStuff164k~\cite{cCocoStuff} dataset. 
    VPNeXt achieves superior segmentation results, particularly in challenging categories.}
    \label{fig:cocostuff_internimage_vs_vpnext}
\end{figure}

\begin{figure}[ht!]
    \centering
    \includegraphics[width=1.0\linewidth]{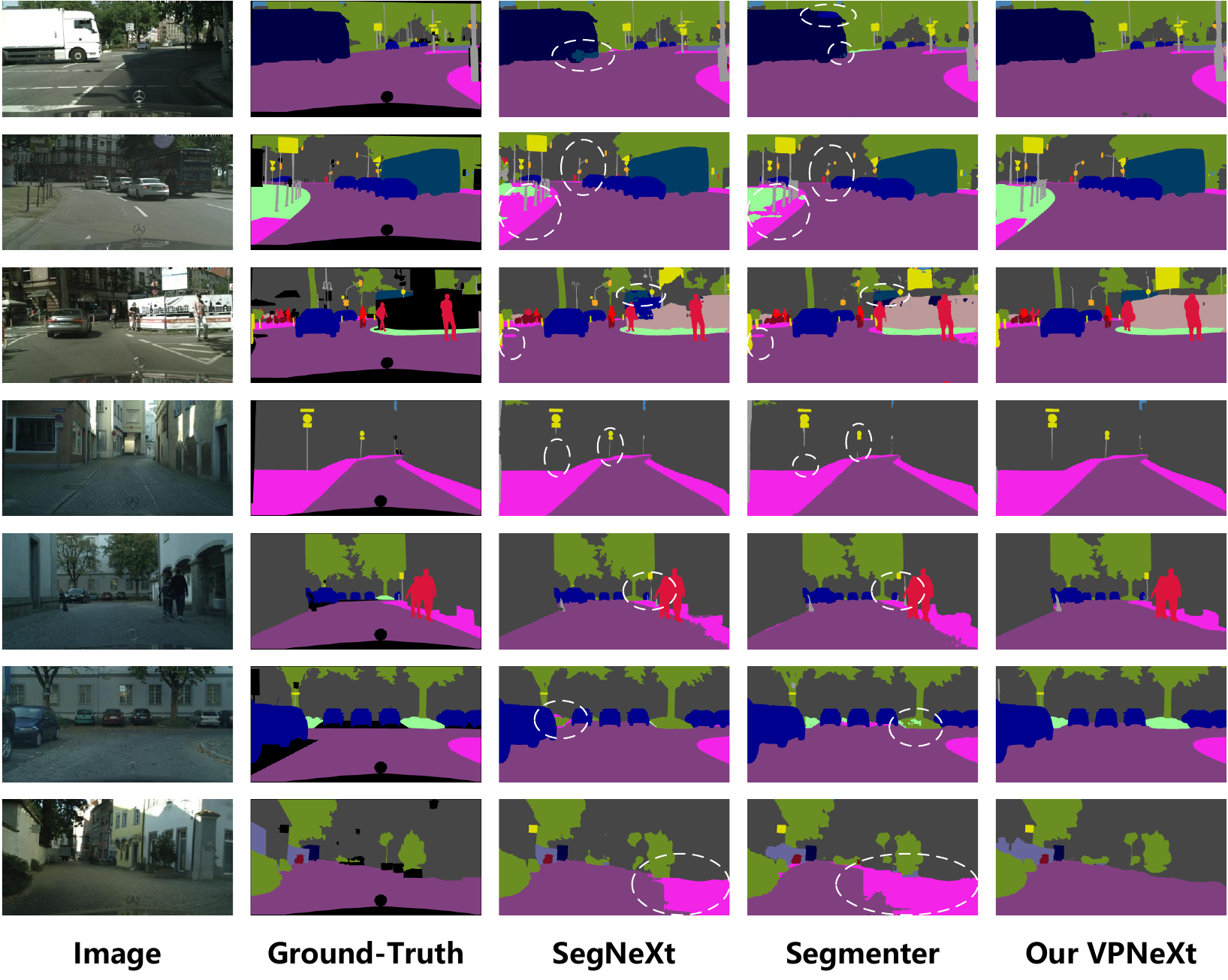}
    \caption{\rv{Visual comparisons between SegNeXt, Segmenter and our proposed VPNeXt on the Cityscapes val set. Zoom in to see better. }
    \rv{We highlight the noticeable shortcomings of the SegNeXt and Segmenter approaches.}}
    \label{fig:ufcn_vs_segnext_segmenter_cityscapes}
\end{figure}

\subsection{Visualization on intermediate feature map}
We first conducted a visualization analysis of the intermediate feature map of the VCR. 
For this analysis, we utilized the feature map from the $15^{\text{th}}$ layer of the ViT. 
As illustrated in Fig.~\ref{fig:vcr_feature_vis}, at the position marked by the red dot (\textcolor{red}{$\bullet$}), the replay-optimized feature map presents significantly stronger and more intensive semantic information regarding intra-class pixels.

\subsection{Visualization on COCOStuff164k dataset}
We \rv{then} conducted a visualization comparison on the challenging COCOStuff164k dataset~\cite{cCocoStuff}. 
As shown in Fig.~\ref{fig:cocostuff_internimage_vs_vpnext}, our proposed VPNeXt achieves significantly better segmentation results compared to the state-of-the-art Mask2Former + InternImage-H~\cite{cMask2Former,cInternImage}, particularly in some challenging categories, such as food.

\subsection{Visualization on Cityscapes dataset}
\rv{
Finally, we provide qualitative visual comparisons among our proposed VPNeXt, SegNeXt, and Segmenter. 
As illustrated in Fig.~\ref{fig:ufcn_vs_segnext_segmenter_cityscapes}, our proposed VPNeXt successfully addresses the upsampling issues of the native ViT and even produces finer details than the pyramid-based SegNeXt.}

\section{Conclusion }
In this paper, we introduce VPNeXt, which includes two innovative modules: VCR and ViTUp. The VCR module effectively aligns intermediate features with high-level features by utilizing both local and global context replay. This approach significantly improves representation accuracy without any runtime overhead. Meanwhile, ViTUp is the first method to reveal hidden pyramid features in Plain ViT, enabling the true upsampling of ViT features using real pyramid features for the first time.
We demonstrated the effectiveness of VPNeXt through extensive experiments, and it also broke the long-standing mIoU wall of the VOC2012 by a large margin, which also stands as the largest improvement since 2015.

\bigskip

\newpage

\bibliographystyle{IEEEtran}
\bibliography{main}

\newpage

\vfill

\end{document}